%% file: main.tex
\documentclass[twoside,11pt]{article}

\input{setup_script}

\usepackage{xspace}
\newcommand{\Cooper}{\texttt{Cooper}\xspace}
\newcommand{\CMP}{\texttt{CMP}\xspace}

\usepackage[nohyperref,preprint]{jmlr2e}

\usepackage{lastpage}
\jmlrheading{??}{2025}{1-\pageref{LastPage}}{6/24; Revised ??/24}{??/24}{21-0000}{Jose Gallego-Posada and Juan Ramirez and Meraj Hashemizadeh and Simon Lacoste-Julien}

\ShortHeadings{\Cooper: A Library for Constrained Optimization in Deep Learning}{Gallego-Posada, Ramirez, Hashemizadeh, Lacoste-Julien}
\firstpageno{1}

\begin{document}

\title{\Cooper: A Library for Constrained Optimization\\ in Deep Learning}

\vspace{-1ex}

\author{
\\
\name Jose Gallego-Posada$^{*,1}$ \email gallegoj@mila.quebec \\
\name Juan Ramirez$^{*,1}$ \email juan.ramirez@mila.quebec \\
\name Meraj Hashemizadeh$^{*,1}$ \email merajhashemi@yahoo.co.uk \\
\name Simon Lacoste-Julien$^{1,2}$ \\
\addr $^1$Mila and DIRO, Université de Montréal, Canada \\
\addr $^2$Canada CIFAR AI Chair
}
\editor{Alexandre Gramfort}

\maketitle

\begin{abstract}%

\hspace{1mm} \Cooper is an open-source package for solving constrained optimization problems involving deep learning models. 
\Cooper implements several Lagrangian-based first-order update schemes,  
making it easy to combine constrained optimization algorithms with high-level features of PyTorch such as automatic differentiation, and specialized deep learning architectures and optimizers. %
Although \Cooper is specifically designed for deep learning applications where gradients are estimated based on mini-batches, it is suitable for general non-convex continuous constrained optimization.
\Cooper's source code is available at \href{https://github.com/cooper-org/cooper}{\texttt{https://github.com/cooper-org/cooper}}.

\end{abstract}


\section{Introduction}
\label{sec:intro}

\vspace{-2ex}

\blfootnote{\hspace{-5ex} $^\star$Equal contribution.}

The rapid advancement and widespread adoption of algorithmic decision systems, such as large-scale machine learning models, have generated significant interest from academic and industrial research organization in enhancing the robustness, safety, and fairness of these systems. These research efforts are typically driven by governmental regulations \citep{EU-AIAct} or ethical considerations \citep{montreal2018}.

The ability to enforce complex behaviors in machine learning models is a central component for ensuring compliance with the mentioned regulatory and ethical guidelines. Constrained optimization offers a rigorous conceptual framework accompanied by algorithmic tools for reliably training machine learning models that satisfy the desired requirements. These requirements can often be formally encoded as numerical (equality or inequality) constraints accompanying the training objective of the model:
\begin{equation}
    \label{eq:cmp_definition}
    \underset{\vx}{\text{min}} \, f(\vx) \hspace{2mm}
    \text{subject to} \hspace{2mm}  \vgx \le \vzero \hspace{2mm}
    \text{and} \hspace{2mm} \vhx = \vzero.
\end{equation}
For example, \citet{dai2024safe} successfully leverage a constrained optimization approach for striking a balance between the helpfulness and harmfulness of large language models trained with reinforcement learning from human feedback \citep{christiano2017deep, ouyang2022training}.
Other works have demonstrated the benefits of constrained optimization techniques in fairness \citep{cotter2019proxy, hashemizadeh2024balancing}, safe reinforcement learning \citep{stooke2020responsive}, active learning \citep{elenter2022lagrangian} and model quantization \citep{hounie2023neural}.
Our previous work has highlighted the tunability advantages of constrained optimization over penalized formulations (where regularizers are incorporated as penalties in the objective function) for training sparse models \citep{gallego2022controlled}

This paper presents \Cooper, a library for solving constrained optimization problems with PyTorch \citep{pytorch}. 
\Cooper aims to facilitate the use of constrained optimization methods in machine learning research and applications.
It implements several first-order update schemes for Lagrangian-based constrained optimization, along with specialized features for tackling problems with large numbers of (possibly non-differentiable) constraints. 
\Cooper benefits from PyTorch for efficient tensor computation and automatic differentiation.

\textbf{Key differentiators.}
\Cooper is a general-purpose library for non-convex constrained optimization, with a strong emphasis on deep learning.
In particular, \Cooper has been designed with native support for the framework of stochastic first-order optimization using mini-batch estimates that is prevalent in the training of deep learning models.

\Cooper's Lagrangian-based approach makes it suitable for a wide range of applications. 
However, some optimization problems enjoy special structure and admit specialized optimization algorithms with enhanced convergence guarantees.
We recommend the use of \Cooper \textit{unless} specialized algorithms are available for a given application.

\textbf{Existing constrained optimization libraries.}
A notable precursor of \Cooper, which is not actively maintained, is TensorFlow's TFCO \citep{cotter2019tfco}. 
We developed \Cooper in response to the shift of the machine learning research community towards PyTorch.
\Cooper is heavily inspired by the design of TFCO.

Among the most popular alternatives for \textit{convex} constrained optimization, we highlight the CVXPY library \citep{diamond2016cvxpy}.
CVXPY provides a modeling language for disciplined convex programming in Python and automates the transformation of the problem into a canonical form, before executing open-source or commercial solvers.
CVXPY is not focused on non-convex problems and thus not suitable for deep learning applications.

CHOP \citep{negiar2020chop} and GeoTorch \citep{lezcano2021geotorch} are alternatives for constrained optimization in PyTorch. JAXopt \citep{blondel2021jaxopt} is a JAX-based option. These libraries rely on the existence of efficient projection operators, linear minimization oracles, or specific manifold structure in the constraints---whereas \Cooper is more generic and does not rely on these specialized structures.

\textbf{Impact.} \Cooper has enabled several papers published at top machine learning conferences: \citet{gallego2022controlled,lachapelle2022partial,ramirez2022l0onie,zhu2023generalized,hashemizadeh2024balancing,sohrabi2024pi,lachapelle2024nonparametric,jang2024active,navarin2024physics,chung2024novel}.

\vspace{-1ex}
\section{Algorithmic overview}
\label{sec:overview}

\textbf{Problem formulation.}
Constrained optimization problems involving the outputs of deep neural networks are typically non-convex. 
A general approach to solving non-convex constrained problems is finding a min-max point of the Lagrangian associated with the constrained optimization problem:
\begin{equation}
    \label{eq:lagrangian_x_lambda_mu}
    \underset{\vx}{\text{min}} \underset{\vlambda \ge \vzero, \, \vmu}{\text{max}}  \Lag(\vx, \vlambda, \vmu) \triangleq f(\vx) + \vlambda\T \vgx + \vmu\T \vhx,
\end{equation}
where $\vlambda \geq \vzero$ and $\vmu$ are the Lagrange multipliers for the inequality and equality constraints, respectively. 
Solving the min-max problem in \cref{eq:lagrangian_x_lambda_mu} is equivalent to the original problem in \cref{eq:cmp_definition}, \emph{even if some of the functions are non-convex}.
We refer the interested reader to the works by \citet{platt1988constrained,boyd2004convex,nocedal2006numerical,bertsekas2016nonlinear} for comprehensive overviews on the theoretical and algorithmic aspects of constrained optimization.

\textbf{Update schemes.}
\Cooper implements several variants of (projected) gradient descent-ascent (GDA) to solve \cref{eq:lagrangian_x_lambda_mu}. 
The simplest approach is simultaneous GDA:
\begin{subequations}
\label{eq:sim_gda}
\begin{align}
    \vx_{t+1} &\leftarrow \texttt{PrimalOptimizerStep}\big(\vx_t, \nabla_{\vx} \Lag (\vx_t, \vlambda_t, \vmu_t)\big), \\
    \vlambda_{t+1} &\leftarrow \left[ \texttt{DualOptimizerStep}\big(\vlambda_t, \vg(\vx_t)\big) \right]_+, \\
    \vmu_{t+1} &\leftarrow \texttt{DualOptimizerStep}\big(\vmu_t, \vh(\vx_t) \big),
\end{align}
\end{subequations}
where $[\, \cdot \,]_+$ is an element-wise projection onto $\reals_{\geq 0}$ to ensure the non-negativity of inequality multipliers. 
Note that the gradients of the Lagrangian with respect to $\vlambda$ and $\vmu$ simplify to $\vg(\vx_t)$ and $\vh(\vx_t)$, respectively.

\textbf{Convergence properties.} 
Recent work demonstrates that GDA can work in practice for Lagrangian constrained optimization \citep{gallego2022controlled,sohrabi2024pi}, although it may diverge for general min-max games \citep{gidel2018variational}.

\textbf{Optimizers.} 
\Cooper allows the use of generic PyTorch optimizers to perform the primal and dual updates in \cref{eq:sim_gda}. This enables reusing pre-existing pipelines for unconstrained minimization when solving constrained optimization problems using \Cooper. 

\textbf{Additional features.}
\Cooper implements the Augmented Lagrangian \citep[\S 5.2.2]{bertsekas2016nonlinear} and Quadratic Penalty \citep[\S 17.1]{nocedal2006numerical} formulations.
\Cooper also implements the proxy-Lagrangian technique of \citet{cotter2019proxy}, which allows for solving constrained optimization problem with \textit{non-differentiable} constraints. 
Moreover, \Cooper supports alternative update schemes to simultaneous GDA such as \textit{alternating} GDA and extragradient \citep{korpelevich1976extragradient,gidel2018variational}. Finally, \Cooper implements the $\nu$PI algorithm \citep{sohrabi2024pi} for improving the multiplier dynamics. 

\section{Using \Cooper}

\begin{figure}[!ht]
    \vspace{-1cm}
    \centering
    \includegraphics[width=0.85\textwidth]{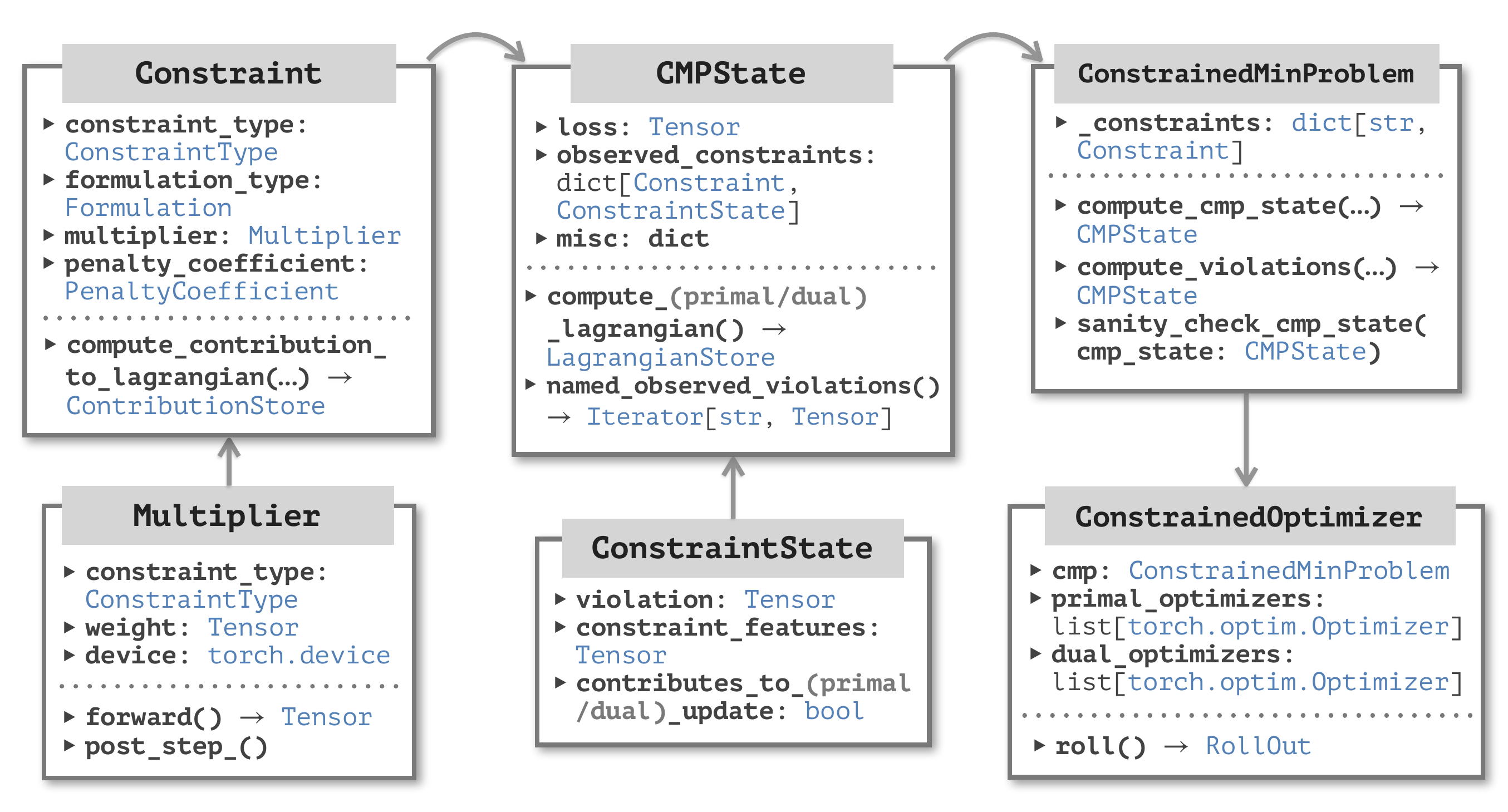}
    \vspace{-2ex}
    \caption{Dependency graph between the main classes in \Cooper's API.}
    \label{fig:structure}
\end{figure}

\Cref{fig:structure} presents \Cooper's main classes. 
The user implements a \texttt{ConstrainedMinimization-}
\texttt{Problem} (\CMP) holding \texttt{Constraint} objects, each in turn holding a corresponding \texttt{Multiplier}.
The \CMP's \texttt{compute\_cmp\_state} method returns the objective value and constraints violations, stored in a \texttt{CMPState} dataclass. 
\texttt{CooperOptimizer}s wrap the primal and dual optimizers and perform updates (such as simultaneous GDA). The \texttt{roll} method of \texttt{CooperOptimizer}s is a convenience function to (i) perform a \texttt{zero\_grad} on all optimizers, (ii) compute the Lagrangian, (iii) call its \texttt{backward} and (iv) perform the primal and dual optimizer steps.

\Cref{algo:quick_start} presents a code example for solving a norm-constrained logistic regression problem with \Cooper. 
This code illustrates the ease of integration of \Cooper with a standard PyTorch training pipeline involving the use of a dataloader, GPU acceleration and the Adam optimizer \citep{kingma2015adam} for the primal parameters.

\section{Software overview}
\label{sec:usage}

\textbf{Installation.}
\Cooper can be installed in Python 3.9-3.11 via \texttt{pip install cooper-optim}. It is supported on Linux, macOS, and Windows and is compatible with PyTorch 1.13-2.3.

\textbf{Collaboration and code quality.} \Cooper is hosted on GitHub under an MIT open-source license.
The library is actively maintained and we welcome external contributions that comply with \Cooper's contribution guide.
We make extensive use of type-hints and use ruff \citep{ruff} as a linter and automatic formatter.
Continuous integration practices are in place to ensure that new contributions pass all tests and comply with the style guidelines before being merged.

All new contributions are expected to be tested following the contribution guidelines. For instance, every optimization scheme counts with both low-level tests ensuring that individual updates are performed correctly, and high-level tests on convex problems checking convergence to verified solutions\footnote{We rely on CVXPY \citep{diamond2016cvxpy} to obtain solutions with optimality certificates.}.
The line coverage of our tests is above $95\%$.

\input{quick_start_code}

\textbf{Documentation.}
\Cooper provides extensive \href{https://cooper.readthedocs.io}{documentation} for all features.
We include formal statements of the updates implemented by all optimizers along with references to relevant sources. We provide quick-start guides aimed at i) users familiar with deep learning problems, and ii) to a broader audience of users interested in generic non-convex constrained optimization problems. Additionally, we have made available several well-documented \href{https://cooper.readthedocs.io/en/latest/auto_tutorials/index.html}{tutorials} illustrating the use of \Cooper's core features.

\section{Conclusion}
\label{sec:conclusion}

\Cooper provides tools for solving constrained optimization problems in PyTorch. 
The library supports several Lagrangian-based first-order update schemes and has been successfully used in machine learning research projects. 
The structure of \Cooper allows for easy implementation of new features such as alternative problem formulations, implicitly parameterized Lagrange multipliers, and additional \texttt{CooperOptimizer} wrappers. 
Implementing a version of \Cooper for JAX \citep{jax2018github} constitutes promising future work.

\acks{
This work was partially supported by an IVADO PhD Excellence Scholarship, the Canada CIFAR AI Chair program (Mila), the NSERC Discovery Grant RGPIN2017-06936 and by Samsung Electronics Co., Ldt. Simon Lacoste-Julien is a CIFAR Associate Fellow in the Learning in Machines \& Brains program.

We would like to thank Manuel Del Verme, Daniel Otero, and Isabel Urrego for useful discussions during the early stages of this work.

Many \Cooper features arose during the development of several research papers. We would like to thank our co-authors Yoshua Bengio, Juan Elenter, Akram Erraqabi, Golnoosh Farnadi, Ignacio Hounie, Alejandro Ribeiro, Rohan Sukumaran, Motahareh Sohrabi and Tianyue (Helen) Zhang.

We thank Sébastien Lachapelle and Lucas Maes for their feedback on this manuscript.
}

\vskip 0.2in
\bibliography{refs}

\end{document}

%% file: setup_script.tex
\usepackage[svgnames,dvipsnames]{xcolor}

\input{math_commands}

\definecolor{linkcolor}{RGB}{0,120,130}

\usepackage[
    colorlinks=true, %
    pdfborder={0 0 0},
    linkcolor=linkcolor,
    citecolor=linkcolor,
    filecolor=linkcolor,
    urlcolor=linkcolor,
    pagebackref=true
]{hyperref} 
\usepackage{url}

\renewcommand*\backref[1]{\ifx#1\relax \else (Cit. on p. #1) \fi}

\usepackage[capitalize]{cleveref}

\usepackage[frozencache,cachedir=.]{minted}
\definecolor{myYellow}{RGB}{245, 238, 26}
\colorlet{ColorHighlight}{myYellow!15}

\newcommand\blfootnote[1]{%
  \begingroup
  \renewcommand\thefootnote{}\footnote{#1}%
  \addtocounter{footnote}{-1}%
  \endgroup
}

%% file: math_commands.tex
\usepackage{amsfonts}
\usepackage{nicefrac}       %
\usepackage{xfrac}
\usepackage{amsmath}
\usepackage{amsthm}
\usepackage{amssymb}
\usepackage{cancel}
\usepackage{mathtools}
\usepackage{bbm}
\usepackage{bm}

\usepackage{cases}         %

\usepackage{mathalpha}

\usepackage{microtype}      %
\usepackage{enumitem}

\usepackage{pifont}

\newcommand{\reals}{\mathbb{R}}

\newcommand{\T}{^{\top}} %

\newcommand{\vzero}{\boldsymbol{0}}

\newcommand{\vg}{\boldsymbol{g}}
\newcommand{\vh}{\boldsymbol{h}}

\newcommand{\vx}{\boldsymbol{x}}

\newcommand{\Lag}{\mathfrak{L}}
\newcommand{\vlambda}{\boldsymbol{\lambda}}
\newcommand{\vmu}{\boldsymbol{\mu}}

\newcommand{\vgx}{\vg(\vx)}
\newcommand{\vhx}{\vh(\vx)}

\definecolor{lightgray}{RGB}{230, 230, 230}
\definecolor{mathred}{RGB}{204, 69, 90}
\definecolor{mathblue}{RGB}{4, 78, 112}
\definecolor{mathgreen}{RGB}{1, 135, 70}

%% file: quick_start_code.tex
\begin{listing}
\vspace*{-9mm}

\begin{minted}
[
frame=lines,
framesep=1mm,
baselinestretch=1,
fontsize=\footnotesize,
linenos,
python3,
highlightlines={7, 11-15, 30, 40-42, 51},
highlightcolor=ColorHighlight
]
{python}
import cooper
import torch

train_loader = ... # Create a PyTorch DataLoader
loss_fn = torch.nn.CrossEntropyLoss()

class NormConstrainedLogisticRegression(cooper.ConstrainedMinimizationProblem):
    def __init__(self, norm_threshold: float):
        self.norm_threshold = norm_threshold
        multiplier = cooper.multipliers.DenseMultiplier(num_constraints=1, device=DEVICE)
        self.norm_constraint = cooper.Constraint(
            multiplier=multiplier,
            constraint_type=cooper.ConstraintType.INEQUALITY,
            formulation_type=cooper.formulations.Lagrangian,
        )

    def compute_cmp_state(self, model, inputs, targets) -> cooper.CMPState:
        logits = model.forward(inputs.view(inputs.shape[0], -1))
        loss = loss_fn(logits, targets)

        sq_l2_norm = model.weight.pow(2).sum() + model.bias.pow(2).sum()
        # Constraint violation uses the convention "g(x) \leq 0"
        norm_constraint_state = cooper.ConstraintState(violation=sq_l2_norm - self.norm_threshold)

        misc = {"batch_accuracy": ...} # useful for storing any additional information

        # Declare observed constraints and their measurements 
        observed_constraints = {self.norm_constraint: norm_constraint_state}

        return cooper.CMPState(loss=loss, observed_constraints=observed_constraints, misc=misc)

cmp = NormConstrainedLogisticRegression(norm_threshold=1.0)

# Create a Logistic Regression model and primal and dual optimizers
model = torch.nn.Linear(in_features=IN_FEATURES, out_features=NUM_CLASSES, bias=True).to(DEVICE)
primal_optimizer = torch.optim.Adam(model.parameters(), lr=1e-3)
# Must set `maximize=True` since the Lagrange multipliers solve a _maximization_ problem
dual_optimizer = torch.optim.SGD(cmp.dual_parameters(), lr=1e-2, maximize=True)

cooper_optimizer = cooper.optim.SimultaneousOptimizer(
    cmp=cmp, primal_optimizers=primal_optimizer, dual_optimizers=dual_optimizer
)

for epoch_num in range(NUM_EPOCHS):
    for inputs, targets in train_loader:
        inputs, targets = inputs.to(DEVICE), targets.to(DEVICE)
        
        # `roll` is a function that packages together the evaluation of the loss, call for 
        # gradient computation, the primal and dual updates and zero_grad 
        compute_cmp_state_kwargs = {"model": model, "inputs": inputs, "targets": targets}
        roll_out = cooper_optimizer.roll(compute_cmp_state_kwargs=compute_cmp_state_kwargs)
        # `roll_out` is a struct containing the loss, last CMPState, and the primal 
        # and dual Lagrangian stores, useful for inspection and logging

torch.save(model.state_dict(), 'model.pt') # Regular  model checkpoint
torch.save(cmp.state_dict(), 'cmp.pt') # Checkpoint for multipliers and penalty coefficients
torch.save(cooper_optimizer.state_dict(), 'cooper_optimizer.pt') # Primal/dual optimizer states
\end{minted}

\vspace{-2ex}
\caption{Example code for solving a norm-constrained logistic regression task using \Cooper.}
\label{algo:quick_start}
\end{listing}